\documentclass[10pt, conference]{IEEEtran}

\usepackage{hyperref}
\usepackage[cmex10]{amsmath}

\usepackage[caption=false,font=footnotesize]{subfig}

\usepackage{multirow}
\usepackage{multicol}

\usepackage{cite}

\usepackage{wrapfig}   
\usepackage{textcomp}  

\usepackage{atbegshi,picture} 
\usepackage{eso-pic, rotating, graphicx}

\AtBeginShipout{\AtBeginShipoutUpperLeft{%
  \put(\dimexpr\paperwidth-1cm\relax,-1cm){\makebox[1pt][r]{{2017 IEEE Congress on Evolutionary Computation (CEC) PREPRINT; the camera ready version will be published in IEEE proceedings}}}%
}}

\AddToShipoutPicture{
  \ifnum\value{page}>1
    \put(20,400){\rotatebox{90}{\scalebox{1.2}{AUTHOR VERSION}}}
   \fi
}

\begin{document}

\title{Tackling Dynamic Vehicle Routing Problem with Time Windows by means of Ant Colony System}

\author{
\IEEEauthorblockN{Raluca Necula}
\IEEEauthorblockA{Faculty of Computer Science\\
Alexandru Ioan Cuza University\\
Iasi, Romania\\
Email: raluca.necula@info.uaic.ro}

\and
\IEEEauthorblockN{Mihaela Breaban}
\IEEEauthorblockA{Faculty of Computer Science\\
Alexandru Ioan Cuza University\\
Iasi, Romania\\
Email: pmihaela@info.uaic.ro}

\and
\IEEEauthorblockN{Madalina Raschip}
\IEEEauthorblockA{Faculty of Computer Science\\
Alexandru Ioan Cuza University\\
Iasi, Romania\\
Email: mionita@info.uaic.ro}
}

\maketitle

\begin{abstract}
The Dynamic Vehicle Routing Problem with Time Windows (DVRPTW) is an extension of the well-known Vehicle Routing Problem (VRP), which takes into account the dynamic nature of the problem. This aspect requires the vehicle routes to be updated in an ongoing manner as new customer requests arrive in the system and must be incorporated into an evolving schedule during the working day. Besides the vehicle capacity constraint involved in the classical VRP, DVRPTW considers in addition time windows, which are able to better capture real-world situations. Despite this, so far, few studies have focused on tackling this problem of greater practical importance. To this end, this study devises for the resolution of DVRPTW, an ant colony optimization based algorithm, which resorts to a joint solution construction mechanism, able to construct in parallel the vehicle routes. This method is coupled with a local search procedure, aimed to further improve the solutions built by ants, and with an insertion heuristics, which tries to reduce the number of vehicles used to service the available customers. The experiments indicate that the proposed algorithm is competitive and effective, and on DVRPTW instances with a higher dynamicity level, it is able to yield better results compared to existing ant-based approaches.
\end{abstract}


\section{Introduction}
The Vehicle Routing Problem plays a central part in capturing logistics and distribution operations, with an impact in the economy, more prominent as the need for transportation is increasingly growing. Thus, this class of problems has a practical importance and has attracted a lot of research interest over the years. Recently, the technological advancements made possible the use of mobile devices to enable the direct communication between the clients and the drivers, such that a driver could dynamically change his plan while executing his route. Also, the emergence of global positioning systems allows a dispatcher to know the current position of a driver and communicate him in a timely manner the next customer to visit on his route. This leads to a VRP variant known as Dynamic Vehicle Routing Problem (DVRP), also referred to as online or real-time VRP.

As indicated in \cite{Braekers2016}, where the authors perform a review on VRP literature,
more challenging problems such as DVRP began being studied more intensively only recently, due to their ability to capture real-world scenarios.
Possible sources of dynamism in DVRP include: customers requests, travel time and demands, which can be dynamically revealed for a set of known customers. In this paper, we considered that the dynamic nature of DVRP is given by the customers requests, which is the most common source of dynamism, as pointed out in \cite{Pillac2013}. Therefore, DVRP allows customers requests to be serviced in a real-time manner, while the drivers have already started their routes, visiting in their way other customers.
Thus, the algorithm needs to adjust the ongoing vehicle routes, such as to take into account these dynamic customer requests, while maintaining the feasibility of solution.

The aim of this study is to design and analyse ant colony optimization based algorithms, that are able to achieve good quality solutions
for DVRPTW instances with a higher level of dynamicity. To this end, we propose in this paper an ant colony system (ACS) based algorithm, relying on a joint mechanism for constructing the solutions, in which the vehicle and the next customer to be added in its tour are simultaneously selected during the transition step. Our approach is hybridized with a local search procedure, consisting of two operators, relocate and exchange, that are applied in an iterative manner, until no further improvement is possible. Moreover, we integrated in our method an insertion heuristics to better incorporate unvisited customers in the existing tours, thus reducing the number of vehicles needed to service all the customers.

The remainder of this paper is structured as follows. Section \ref{sec:DVRPTW} describes the static variant of the considered DVRPTW problem. Section \ref{sec:relatedWork} presents a survey on existing literature related to DVRPTW. In Section \ref{sec:ACS} the ant colony system optimization algorithm is summarized, as the underlying metaheuristic for our approach that will be described in Section \ref{sec:investigated algorithm}. In Section \ref{sec:experiments} the experimental study and the obtained results will be presented. The paper concludes with Section \ref{sec:conclusions}, which draws the final remarks and offers future research directions.

\section{The vehicle routing problem with time windows} \label{sec:DVRPTW}
In this section, we describe the (static) vehicle routing problem with time windows (VRPTW), which is the static variant of our considered DVRPTW problem.

As in the classical VRP or capacitated VRP, VRPTW requires to find a set of routes for each vehicle, such that each customer, having a given demand, is visited exactly once by a single vehicle and the total demand on the route does not exceed the vehicle's capacity.
The fleet of vehicles is assumed to be homogeneous, all the vehicles are located at a single depot, where they start and end their route, and each vehicle has a limited capacity. Besides this, VRPTW brings the additional constraint of time windows associated to a customer, such that a customer $i$ must be serviced by a vehicle within a time interval, defined as $[e_i, l_i]$, where $e_i$ is the earliest arrival time, and $l_i$ is the latest arrival time. The service at customer $i$ cannot start before its earliest arrival time, and if the vehicle arrives at the customer earlier than $e_i$, waiting occurs. Also, the service at each customer takes an amount of time, denoted by $s_i$, required for the pickup and/or delivery of goods or services.
In case of hard time windows constraints, the vehicle must arrive at the customers not later than the latest arrival time, constraint that must be satisfied for each customer within a vehicle's tour in order to maintain the solution feasible. There is also the possibility of applying penalties if the services start after the allowed time windows, which is the case of soft time windows. The depot's time window induces a constraint regarding the maximum total route time, meaning that each vehicle route must start and end within the time window associated with the depot.

In this study, we consider hard time windows constraints and the existence of a homogeneous fleet of vehicles, that are located at a single depot. Also, we assume that the primary objective of VRPTW is to minimize the number of tours (vehicles), whilst a second objective is to minimize the total traveled distance needed to supply all customers in their required hours. This means that the number of tours minimization takes precedence over total distance minimization. In case of solutions with the same number of vehicles, those solutions of lower distance are preferred.

By its description, it can be noticed that VRPTW is more complicated than multiple traveling salesman problem (mTSP), problem that we have tackled in our previous studies \cite{HAIS2015,ICTAI2015} with ant colony based methods, and for which we proposed a benchmark\footnote{\url{http://profs.info.uaic.ro/~mtsplib/}}. Thus, mTSP can be seen as a relaxation of VRPTW, if removing the time windows associated with the customers, and assuming that vehicles have unlimited capacity, which imposes no restriction on the number of customers to be visited on a vehicle's route.

\section{Related work}\label{sec:relatedWork}

During the last decades, the VRP and his famous VRPTW extension have been an intensive research area. Heuristic and exact optimization approaches have been developed for the VRPTW problem. Surveys of proposed techniques can be found in \cite{Cordeau2001,Gendreau2005}.
As the VRP and VRPTW problems are both NP-hard and they generalize the traveling salesman problem (TSP), ant-based methods were proposed for solving them. A multi-ant colony system consisting in two colonies, one for optimizing the number of vehicles and one for optimizing the total travel time, was proposed for the VRPTW problem in \cite{Gambardella1999}.

Because of the recent technological advances, the research on dynamic routing increased. Techniques ranging from linear programming to metaheuristics were designed to address dynamism and uncertainty. A comprehensive review of applications and approaches for dynamic vehicle routing problems is given in \cite{Pillac2013}.
A taxonomy of papers concerning DVRP problems was developed in \cite{Psaraftis2016}.
Also, the interested reader is referred to \cite{Ilin2015}, for a survey of hybrid artificial intelligence algorithms, employed for solving DVRP problems.

Very few previous studies on using metaheuristics on DVRPTW exist, that take into account both the dynamic nature of VRP and the time windows as additional constraint. A first such study is due to \cite{Gendreau1999}, which presents a tabu search algorithm that works on the DVRPTW, but with soft time windows. A hybrid genetic approach based on two populations of individuals which evolve concurrently are used to minimize the total traveled distance, and respectively, the number of unserviced customers, total lateness at customer’s locations and temporal constraint violations in \cite{Alvarenga2005}. A metaheuristic algorithm is developed in \cite{deArmas2015} for tackling two variants of DVRPTW, which consider several real-world constraints: multiple time windows, customers priorities and vehicle-customer constraints. In \cite{Hong2012}, the DVRPTW problem is decomposed into several static VRPTW problems, which are then solved using an improved large neighborhood search algorithm.

To the best of our knowledge, the study from \cite{Veen2013} is the only ant colony based approach found in the literature, that tackles the DVRPTW problem. To this end, the authors propose an ACO based method, denoted as \emph{MACS-DVRPTW}, by extending to the dynamical case the state-of-the-art ant algorithm \cite{Gambardella1999} developed for VRPTW. This study was continued in \cite{Yang2016}, where \emph{MACS-DVRPTW} was applied to schedule the routes of a fleet of cars for a surveillance company. Recently, \emph{MACS-DVRPTW} was enhanced in \cite{Yang2015} with two strategies for dealing with DVRPTW problem in which customers have different priority levels.

\section{The Ant colony system} \label{sec:ACS}
Ant Colony Optimization (ACO) is a nature-inspired metaheuristic, which follows the metaphor of real ants that succeed in finding the shortest paths between their nest and food sources. In case of ACO, each artificial ant, referred henceforth as "ant", constructs solutions in a probabilistic manner by iteratively adding components, until a complete solution to the problem is obtained. When building a solution, an ant is influenced by two factors: 1) pheromone trails, updated dynamically at runtime, that reflect the gained search experience of ants, and 2) the heuristic information, which is a static value, dependent upon the problem.

The algorithm presented in this study is based on the Ant Colony System (ACS) \cite{ACS}, which is a successor of the Ant System (AS) \cite{AS}.
AS is the first ACO algorithm proposed in the literature, that was initially designed for solving instances of TSP.
ACS is an improvement over AS and it brings the following changes: 1) a modified state transition rule, which is more focused on exploitation, 2) the inclusion of a local pheromone update, which reduces the amount of pheromone on edges visited by ants, in order to increase exploration and prevent an early stagnation of search, and 3) the global pheromone update is performed by a single ant, generally being the best so far ant, which produced the best solution.

\section{The algorithm investigated for DVRPTW} \label{sec:investigated algorithm}
This section describes our approach designed for the DVRPTW problem, which is based on the ant colony system metaheuristic, outlined in Section \ref{sec:ACS}. To tackle the dynamic nature of the problem, we adopted in our algorithm the time slices and nodes commitment approach from the \emph{MACS-DVRPTW} ant based solver, introduced in \cite{Veen2013}. Accordingly, a working day of $t_{wd}$ seconds is equally split into $n_{ts}$ time slices, the length of a time slice ($t_{ts}$) being computed as: $t_{ts} = t_{wd}/n_{ts}$. Thus, the initial DVRPTW problem is divided into $n_{ts}$ static problems, that will be solved consecutively. In this context, we introduce the concept of current problem configuration, which denotes one of these static problems, defined by the subset of available nodes (customer requests). We will denote by $S^{best}$ the currently optimal solution, which is the best feasible solution including all the available customer requests known so far.

Our approach, referred henceforth as \emph{DVRPTW-ACS}, is composed of two parts, that will be presented in the next subsections, and each part is running in a separate thread. The core part of the algorithm, briefly denoted as the "\emph{planner}", is concerned with the time slices management and with the dynamic nature of the problem. The other part is responsible with the optimization task, by running the ACS based algorithm on the current problem configuration.

\subsection{The planner} \label{sec:planner}
Before the beginning of the working day, an initial feasible solution (with respect to capacity, time window and vehicle arrival time at the depot constraints), is built using the time-oriented nearest neighbour heuristic, introduced in \cite{Solomon1987}.
The time-oriented nearest neighbour heuristic is a sequential tour building algorithm, that takes into account both geographical and temporal closeness of customers. This initial solution defines the tentative tours for the vehicles, considering only a priori nodes, known in the system from the beginning.

Initially, a vehicle starts its tour from the depot, then iteratively adds the closest (in terms of a metric $m$) unvisited customer, relative to the last customer of this tour, while keeping feasible the tour under construction. The metric $m_{ij}$, which defines the closeness between $i$, the last customer on the current emerging tour, and $j$, that refers to any unrouted customer that could be visited next, is defined as:
\begin{equation}
m_{ij}=0.4 \cdot d_{ij} + 0.4 \cdot T_{ij} + 0.2 \cdot u_{ij}
\label{eq:c metric}
\end{equation}
In the previous equation, $d_{ij}$ is the distance between the two customers, $T_{ij}$ means the time difference between the beginning of service at customer $j$ and the completion of service at customer $i$, and is expressed as: $T_{ij}=b_j - (b_i + s_i)$, and $u_{ij}$ represents the urgency of delivery to customer $j$, defined as: $u_{ij}=l_j-(b_i + s_i + d_{ij})$, where $b_j$ denotes the beginning of service at customer $j$, being computed as: $b_j=max(e_j, b_i + s_i + d_{ij})$, assuming that the vehicle travels to the next customer $j$, as soon as it has finished the service at the current customer $i$.

Each time when there are still remaining unrouted customers, that cannot be feasibly appended to the current tour, a new tour, that starts from the depot, is added to incorporate them.

The obtained solution is used to initialize the best so far solution, $S^{best}$, and to compute the value of the initial pheromone level, $\tau_0$, as:
\begin{equation}
\tau_0=(n_{av}\cdot L_{NN})^{-1}
\label{eq:initial_pheromone}
\end{equation}
where $n_{av}$ is the number of available nodes (customers) and $L_{NN}$ denotes the total traveled distance of this solution.

When the working day begins, a timer is started which keeps track of the current time, so that when the working day is over, the algorithm will stop its execution. Also, the planner starts the ant colony thread, which will perform its optimization task, considering only the currently available customers.

Taking into account the current time, the planner repeatedly verifies whether a new time slice began. If this is the case, the planner checks: 1) if there are new nodes that became available in the meantime, during the last time slice, and 2) if there are new nodes that must be committed. If one of these two conditions is met, the planner stops the execution of the ACS based algorithm, since the problem definition has changed.

The next node $i$, after the last committed node from a tour of the best so far solution, $S^{best}$, is marked as committed if the following condition holds: $b_i \leq idx_{ts} \cdot t_{ts}$, where $idx_{ts}$ is the index of the current time slice, and the $idx_{ts} \cdot t_{ts}$ product denotes the time when the current time slice ends.
When a node becomes committed, its location within the vehicle's tour cannot be changed anymore and it will be found at this position in its tour in the final solution to be returned by the algorithm at the end of the working day.

After committing the necessary nodes from the tours of $S^{best}$, if condition 1) holds, the list of available customer requests is updated and the new available nodes are added in the tours of $S^{best}$ by adopting the insertion technique from \emph{I1} insertion heuristic \cite{Solomon1987}. The insertion heuristic starts by computing for each unrouted node its best feasible insertion place in the tours of $S^{best}$ according to a criterion ($c_1$). The score of inserting a customer $u$ between two adjacent customers $i$ and $j$ on a route, is computed as: $c_1(i, u, j)=0.1 \cdot c_{11}(i, u, j)+ 0.9 \cdot c_{12}(i, u, j)$, where $c_{11}(i, u, j)=d_{iu}+d_{uj}-d_{ij}$ and $c_{12}(i, u, j)=b_{j_u}-b_j$, where $b_{j_u}$ denotes the new time of beginning the service at customer $j$, after inserting customer $u$ in the route. For a given node, its best insertion position is the one that achieves the minimum score for the $c_1$ criterion. Based on this, the best customer $u$ to be inserted in the route is selected to be the one that obtains the minimum value for $c_2$ criterion, defined as: $c_2(i, u, j)=2.0 \cdot d_{0u} - c_1(i, u, j)$, where $d_{0u}$ denotes the distance between the depot and customer $u$.
The insertion heuristic is applied in an iterated manner, until no further feasible insertions are possible.

If there are still remaining available nodes which couldn't be feasibly inserted in the existing tours of $S^{best}$, a new tour is added, which starts from the depot. The unrouted nodes will be tried to be added in this tour by following the same construction mechanism used in the time-oriented nearest neighbour heuristic.
This process continues until all the available customer requests are incorporated in $S^{best}$, meaning that a complete feasible solution was obtained.

After the necessary commitments were made and the new available nodes were added in the best so far solution, the ant colony is restarted, performing its optimization task on the new problem configuration. The algorithm stops after $t_{wd}$ seconds of execution, which marks the ending of the working day. At this point in time, the best solution obtained so far is returned as the output of \emph{DVRPTW-ACS}, indicating the final tours to be traveled by each vehicle.

\subsection{The ACS based algorithm} \label{sec:ACS_algorithm}
Our \emph{DVRPTW-ACS} approach relies on the ACS algorithm for performing the optimization task, such that all the available customers are visited only once, while minimizing the number of vehicles and the total traveled distance. An ant's solution is composed of a set of tours, each one designating the route to be traveled by a vehicle. As mentioned in Section \ref{sec:DVRPTW}, the formulation of the considered VRPTW problem assumes a hierarchy between these two objectives, namely the minimization of the number of vehicles takes precedence over the minimization of the traveled distance. This implies that a solution which uses less vehicles will be preferred over a solution with more vehicles, even if it has a smaller traveled distance.

Unlike the \emph{MACS-DVRPTW} solver which resorts to two colonies, \emph{ACS-VEI} and \emph{ACS-TIME}, each one trying to optimize a different objective, our algorithm uses one single colony, which simultaneously optimizes the two objectives, the number of vehicles and the total traveled distance.
Consequently, \emph{MACS-DVRPTW} uses two pheromone trail matrices, one for each colony, whereas our approach involves only one pheromone trial matrix. Besides that, \emph{DVRPTW-ACS} works only with feasible solutions, in which all the known customers are incorporated, which is different from \emph{MACS-DVRPTW}, where the \emph{ACS-VEI} colony works also with unfeasible solutions.

At the first run of the ACS based algorithm, the pheromone trail matrix is initialized with $\tau_0$, computed as in equation (\ref{eq:initial_pheromone}). Then, at subsequent runs, when the ant colony is restarted, $\tau_0$ is computed like in equation (\ref{eq:initial_pheromone}), except for the fact that $L_{NN}$ is replaced with $L_{best}$, which denotes the total distance of the tours from $S^{best}$. In addition to this, we have adopted the pheromone preservation strategy, employed also in one of the variants of \emph{MACS-DVRPTW}, which allows that a certain amount of pheromone from the previous run of the ant colony to be preserved in the current execution.

In the implementation of the algorithm, we resorted to candidate lists, as indicated in \cite{Gambardella1996}, when solving instances of TSP with ACS. A candidate list of size $cl$ contains for a given node, the most closest $cl$ neighbours, arranged in increasing order according to their distance from the considered node. In our case, all the nodes part of the DVRPTW instance will be considered when computing the candidate list of a node. Then, in the transition step of ACS, when an ant located in node $r$ has to decide the next node to move to, it will first examine the candidate list of node $r$, choosing only those nodes that are revealed. Only when all the available nodes from the candidate list have already been visited, the search of the next customer is extended on the rest of available nodes.

Initially, a vehicle is placed at the depot, where it starts and ends its tour. In an iteration of the ACS based algorithm, several ants build feasible solutions for the problem, that integrate all the available customers and do not violate any of the VRPTW constraints: capacity, time window and vehicle arrival time at the depot. Later on during the algorithm execution, when the planner starts to commit nodes, the committed parts of $S^{best}$ are used to initialize the ants' solutions, being copied at exactly the same positions. Therefore, it may result that an ant's solution to be comprised from the beginning of more than one tour and use more than one vehicle. In case there are committed nodes, the solution construction will continue from these added portions of tours, which will be next extended by ants during the transition step.

Unlike the \emph{MACS-DVRPTW} solver, in which the tours are constructed sequentially and filled up with customers one by one, in our approach the tours are built simultaneously and the vehicles compete towards extending their partially tours with a new available customer. In addition to this, we employed a joint mechanism for building the vehicles tours, in which the vehicle and the customer to be added in its tour, are selected at the same time, during the state transition rule from the standard ACS.
Both the vehicle ($v$) and the next customer to visit ($s$) are chosen by iterating over the existing tours (vehicles) from the set $T$ and the candidate set $C$, comprised of available customers, that have not been visited and committed yet, and can be appended to the tour of $v$ vehicle, while keeping the solution feasible:
\begin{equation}
(v,s) = \begin{cases}
       \arg \max\limits_{v\in T s \in C}({\tau^{\alpha}_{rs}\cdot \eta^{\beta}_{rs})}, & \text{if $rand(0,1) < q_0$}\\
       S, & \text{otherwise}
    \end{cases}
    \label{eq:p_node_s}
\end{equation}
where $r$ represents the ant's current position, being the last customer on the tour denoted by $v$, $\tau_{rs}$ is the pheromone trail associated to edge $rs$, $\eta_{rs}$ is the heuristic information corresponding to edge $rs$ and is computed as: $\eta^{\beta}_{rs} = (1/m_{rs})^{\beta}$, where $m_{rs}$ is defined as in equation (\ref{eq:c metric}).
In addition, $\alpha$ and $\beta$ are parameters that reflect the importance of pheromone, respectively heuristic information, $q_0 \in [0,1]$ is a parameter and
$S$ is a random variable selected according to the following probability distribution:
\[p_{rs} = \left\{
\begin{array}{l l}
\frac{\tau^{\alpha}_{rs}\cdot \eta^{\beta}_{rs}}{\sum_{w\in T}\sum_{u\in C}\tau^{\alpha}_{ru}\cdot \eta^{\beta}_{ru}}, & \text{if $s \in C$}\\
0, & \text{otherwise}
\end{array} \right.\]

During the building process of an ant's solution, when there are remaining unrouted available customers that cannot be feasibly added in the existing tours and the number of these customers is less or equal to ten, we apply the same insertion heuristic used by the planner to update $S^{best}$ with the new available nodes. Its aim is to insert in the existing tours the remaining available unvisited nodes. In this way, the insertion heuristic is applied in order to prevent the addition of new tours needed to incorporate these nodes, thus reducing the number of vehicles used to service all the known customers. After this step, if there are still remaining unvisited customers, that are available, a new tour is added to incorporate them. This process is repeated until the ant's solution covers all the available customers, meaning that a feasible complete solution for the DVRPTW problem was obtained.

Once an ant moves from customer $r$ to $s$, it updates the pheromone level on the edge associated to it, by applying the local pheromone update rule, as in the standard ACS:
\begin{equation}
\tau_{rs}=(1-\rho)\cdot \tau_{rs}+\rho\cdot \Delta \tau_{rs}
\label{eq:local_pheromone}
\end{equation}
where $\rho \in (0,1)$ is a local pheromone decay parameter and for the $\Delta \tau_{rs}$ we have used the following setting, as indicated in \cite{ACS}: $\Delta \tau_{rs} = \tau_0$, where $\tau_0$ is the initial pheromone level, computed as in equation (\ref{eq:initial_pheromone}). The aim of the local pheromone update is to make visited edges less desirable for subsequent ants, enforcing diversity within the same iteration and thus avoiding the situation in which all the ants construct very similar tours.

After all the ants finished to construct their solution, the best iteration ant is selected as the one that built the best solution, either regarding the number of vehicles or the total distance objective. Since the number of vehicles objective has priority over the total distance objective, a solution which uses less vehicles is always preferred over a solution with more vehicles, even if it has a smaller total distance. The global best ant, which corresponds to the best so far solution, is updated, if necessary, based on the best iteration ant. Then it follows the global pheromone update and like in the standard ACS, it reinforces with additional pheromone the edges used by the global best ant:
\begin{equation}
\tau_{rs}=(1-\rho)\cdot \tau_{rs}+\rho\cdot \Delta \tau_{rs}
\label{eq:global_pheromone}
\end{equation}
where
\[ \Delta \tau_{rs} = \left\{
\begin{array}{l l}
(L_{gb})^{-1} , & \text{if $rs \in $ best so far solution}\\
0, & \text{otherwise}
\end{array} \right.\]
where $\rho \in (0,1)$ is the pheromone decay parameter and $L_{gb}$ denotes the total traveled distance of the best so far solution, obtained by the global best ant.

Since it is known from the literature that combining an ACO algorithm with a local search phase can greatly improve its performance, we integrated in \emph{DVRPTW-ACS} a local search procedure. This method consists of two multi-route operators, relocate and exchange \cite{Savelsbergh1992}, which modify simultaneously two different tours. The relocate operator removes one node from a tour and reinserts it into another tour, whilst the exchange operator swaps the position of two nodes from two different tours.
These two operators are applied in an iterated manner, meaning they are called repeatedly, until no further improvement, regarding the number of vehicles or the total distance, can be obtained. We applied the local search procedure to the initial solution, produced by the time-oriented nearest neighbour heuristic, and to the best solution obtained in the current iteration (best iteration ant) of the algorithm.

According to the concept of commitment, which involves that the position of committed nodes within a tour is fixed, the insertion heuristic and the local search operators are applied only to positions that do not belong to the committed parts of a tour. In case of insertion heuristic, this means that a node cannot be inserted in front of a node that is committed, whilst in case of the local search procedure, the positions eligible for relocation and exchange will start from the first uncommitted node within a tour.

The ACS based algorithm iteratively performs its optimization task, trying to improve $S^{best}$ in any of the two objectives (number of vehicles and total distance), until it is stopped by the planner, either because new nodes became available or because there are nodes that must be committed.

\section{Experiments} \label{sec:experiments}

\subsection{Problem instances} \label{sec:instances}
The experimental analysis in this paper is conducted on several DVRPTW instances, belonging to the benchmark introduced in \cite{Veen2013}. These problem instances were constructed starting from the Solomon's 100 customers static VRPTW benchmark\footnote{\url{http://web.cba.neu.edu/~msolomon/problems.htm}}, that was extended to the dynamical case, by specifying a dynamicity level of $X\%$. To this end, the Solomon's benchmark was enhanced with an additional attribute, namely the available time, which conveys the time when a customer's request is revealed to the system. Starting with this time, the customer request is considered to be known and will be taken into account by the algorithm.

A dynamicity of $X\%$ means that a percentage of $X\%$ customer requests have a non-zero available time, these being dynamic requests which are revealed during the working day. The a-priori nodes (customer requests), that are known from the beginning of the algorithm execution, have their available time equal to 0.
The problem instances from the DVRPTW benchmark have specified different values for the dynamicity level, ranging in the interval $[0\%..100\%]$, with an increment of 10\%. The instances with 0\% dynamicity are static VRPTW problems taken from the Solomon's benchmark, whilst the instances with 100\% dynamicity are the most dynamic ones.

The DVRPTW benchmark encompasses euclidian test problems, divided in six categories: R1, R2, C1, C2, RC1 and RC2. The test instances in R1 and R2 categories have nodes with randomly generated coordinates, the nodes from C1 and C2 categories are clustered, whilst in RC1 and RC2 there is a combination of randomly generated and clustered nodes. Test problems of type 1 have a short scheduling horizon, allowing only a few nodes to be visited on a vehicle's route. On the other hand, problems of type 2 have a long scheduling horizon, such that more customers can be serviced by the same vehicle, and thus fewer vehicles are needed to visit all the customers. From each of these six categories, we have chosen two problem instances and for each of it we have selected three different values for dynamicity: 10\%, 50\% and 100\%, resulting in a total of 36 DVRPTW instances, that were used in our experimental study. We have chosen different dynamicity degrees in order to analyze the impact of the dynamicity on the quality of solutions produced by the investigated algorithms.
In each DVRPTW instance, the depot is considered to be the first entry in the list of given customers. Also, the fleet of vehicles is homogeneous and the travel time between two customers is assumed to be equal with the distance between them.

The VRPTW problem instances from the Solomon's benchmark define a time window for each node, including the depot, namely $[e_0, l_0]$, which denotes the scheduling horizon.
At the same time, the specifications of the DVRPTW problem involve a certain length of the working day, 
which establishes when the algorithm will stop from its execution. To comply with these two requirements, we scaled in our algorithm all time related values, as it was done in the \emph{MACS-DVRPTW} solver. More precisely, for each customer request from a particular DVRPTW instance, the values for time windows, service time, available time and distance between two customers, are multiplied with the $s_v$ factor, computed as: $s_v = t_{wd}/(l_0 - e_0)$. In a similar manner, the results reported in Section \ref{sec:results} for the total traveled distance ($TD$) are obtained as: $TD = TD / s_v$.

To allow the reproduction of results, the DVRPTW test instances, used in the experimental part of this study, and the Java source code of our \emph{DVRPTW-ACS} algorithm are available online\footnote{\url{http://profs.info.uaic.ro/~mtsplib/vrp-extension/dvrptw/}}.

\subsection{Parameters setup} \label{sec:parameters}
To allow a fair comparison of the two investigated algorithms, \emph{DVRPTW-ACS} and \emph{MACS-DVRPTW}, we used the same setting of the parameters, their values being taken from \cite{Veen2013}. For the ACS based algorithms we have set: $q_0 = 0.9$, $\alpha = 1.0$, $\beta = 1.0$, $\rho = 0.9$ and the number of ants was set to 10, whilst for the dynamic scenario we have set: $t_{wd}=100$ seconds and $n_{ts}=50$. Also, we have employed in both algorithms the pheromone preservation strategy, by specifying $\rho = 0.3$ in the equation: $\tau_{ij}=(1-\rho)\cdot \tau^{old}_{ij}+\rho\cdot\tau_0$. Besides that, the size of the candidate list, used in our algorithm, was set to $cl=20$.
For both algorithms 30 runs were carried out on a Linux server with 6 GB RAM memory, processor Intel Core 2 Duo P9xxx (Penryn Class Core 2) at 2.5 GHz.

\subsection{Results} \label{sec:results}
To assess the performance of the proposed algorithm, we compared it with another ACO based approach from the literature, namely \emph{MACS-DVRPTW}, described in \cite{Veen2013}. More precisely, among the four variants of ACO based algorithms employed for the \emph{MACS-DVRPTW} solver, we have chosen the ACS based algorithm with pheromone preservation (WPP) and which uses no improved initial solution (IIS), constructed before the start of the working day. We selected this variant of algorithm, denoted as "DVRP, 0.3 wpp, no IIS" in \cite{Veen2013}, since according to the results the authors reported in their study, this variant achieves overall one of the best results on DVRPTW instances with 50\% dynamicity. For performing the comparison of the investigated approaches, we stored for each method the best so far solution and the number of feasible solutions, computed during the working day.

Based on the 30 best so far solutions achieved at the end of each algorithm execution on all DVRPTW instances, we report in Tables \ref{tab:R1results} to \ref{tab:RC2results} the average, minimum (best), maximum (worst) and standard deviation values. These values are computed in terms of the two objectives: number of vehicles (abbreviated in tables as NV) and total traveled distance (abbreviated in tables as TD).

In addition to this, we also indicate the results obtained by two approaches after performing 30 runs on static VRPTW instances. The static VRPTW instances correspond to the considered DVRPTW instances, in which all the customer requests are known beforehand and there is no node to be revealed dynamically during the working day. The instances from the DVRPTW benchmark were constructed such as to still allow obtaining the optimal solution of the corresponding static VRPTW instance. Therefore, the solutions produced for the static instances will act as a reference point when assessing the quality of solutions found by the investigated approaches on dynamic instances. In this sense, we have computed the decline in the solution quality, relative to the solution obtained for the static instance, that will allow us to analyze how much the solution deteriorates with the increase in the dynamicity level. The decline measure is indicated in tables as "increase(\%)" and its value is computed for each of the two objectives as: $increase_k = (minD_k - minS_k)/minS_k \cdot 100$, $k=1, 2$, where $minD_k$ is the minimum value
obtained on the dynamic instance for the k-th objective, and $minS_k$ is the minimum value
obtained on the static instance for the k-th objective.

In the evaluation of the two algorithms, we take into account the hierarchy between the two objectives, meaning that we aim with priority minimum values for the number of vehicles, then secondly we seek for smaller total distance values. In each table, we highlighted with boldface the best (minimum) value from a row, except for the instances where the statistical test did not show statistically significant differences. The end part from the name of a DVRPTW instance reflects its associated dynamicity, such that "0.0" denotes static instances having 0\% dynamicity, "0.1" is for instances with a low dynamicity of 10\%, "0.5" is for instances with 50\% dynamicity, and "1.0" denotes instances having the maximum dynamicity of 100\%.

Analyzing the best values obtained for the two objectives, it can be stated that \emph{MACS-DVRPTW} is a clear winner on most of the static instances, except for \emph{C201-0.0} and \emph{C202-0.0} instances, where our algorithm has a better average performance. Actually, on these instances, the minimum values produced on \emph{C201-0.0} by both algorithms, and on \emph{C202-0.0} by \emph{DVRPTW-ACS}, coincide with the best known solution reported in the literature\footnote{\url{https://goo.gl/YbMGJx}}. Also, on most of DVRPTW instances with a low dynamicity of 10\% and in case of a few instances with higher dynamicity, such as \emph{R103-0.5}, \emph{R104-0.5} and \emph{R104-1.0}, \emph{MACS-DVRPTW} achieves better results than our approach on both objectives. The advantage of our algorithm over \emph{MACS-DVRPTW} becomes apparent as the dynamicity level of instances increases. Thus, on almost all instances of 50\% and 100\% dynamicity, \emph{DVRPTW-ACS} manages to obtain superior solutions with less vehicles and of lower total distance than \emph{MACS-DVRPTW}. The better performance of our algorithm is more noticeable on \emph{C201-0.5}, \emph{C202-0.5}, \emph{C101-1.0}, \emph{C102-1.0} and \emph{C201-1.0} instances, where the values attained by \emph{DVRPTW-ACS} on both objectives are almost half the ones obtained by \emph{MACS-DVRPTW}.

The values recorded by the two approaches for the increase measure reveal the same good performance of \emph{DVRPTW-ACS}. More precisely, it can be noticed that the solution quality on DVRPTW instances with higher dynamicity of 50\% and 100\% is not influenced so much by the increase in the dynamicity. Besides that, on instances with clustered nodes such as \emph{C101-0.5}, \emph{C101-1.0}, \emph{C201-0.5} and \emph{C201-1.0}, \emph{DVRPTW-ACS} is still able to attain the same good results as obtained for the corresponding static instances, as indicated by the zero values for the increase measure. In contrast to this, \emph{MACS-DVRPTW} yields on most instances higher values for the increase measure. Also on some instances with higher dynamicity such as \emph{R201-1.0}, \emph{C201-0.5}, \emph{C201-1.0}, \emph{C202-1.0} and \emph{RC202-1.0}, the value for increase measure in terms of the number of vehicles is greater than or equal to 100, which shows that the decline in the solution quality is significant. Also, compared to the average number of vehicles required on static instances, \emph{MACS-DVRPTW} uses in addition up to 2 vehicles on instances of 10\% dynamicity, on instances of 50\% dynamicity it uses at most extra 6 vehicles, and on instances of 100\% dynamicity it uses at most extra 10 vehicles. On the other hand, our approach requires on average on all DVRPTW instances, regardless of their dynamicity level, up to 2 extra vehicles.
This aspect shows that our algorithm manages to cope better on instances of higher dynamicity, in which fewer customers are known in advance and more nodes are revealed in an ongoing fashion during the working day.

\begin{figure}[h]
	\centering
	\includegraphics[width = 4.4cm]{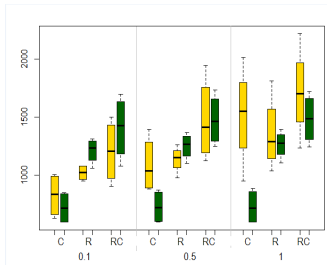}
	\caption{Distribution of total distance objective for dynamicity levels ranging in \{0.1, 0.5, 1\}; yellow is for \emph{MACS-DVRPTW}, green is for \emph{DVRPTW-ACS}. Results are reported for each of the three problem classes (C, R, RC)}
	\label{fig:boxplot}
\end{figure}

Figure \ref{fig:boxplot} illustrates the distribution of the results of the two algorithms with respect to the total distance objective. Yellow boxplots correspond to the results obtained with \emph{MACS-DVRPTW}, and green ones to \emph{DVRPTW-ACS}. The results are grouped based on the instance category and dynamicity level. Analyzing the boxplots, it can be noticed that our algorithm wins on instances from C category for all the dynamicity levels. Also, on R and RC instances with low dynamicity, \emph{DVRPTW-ACS} is surpassed. However, on the same R and RC instances having 100\% dynamicity, \emph{DVRPTW-ACS} outperforms \emph{MACS-DVRPTW}. Besides that, the boxplots reveal that the variance of \emph{DVRPTW-ACS} is smaller, which indicates a more stable algorithm.

We conducted two statistical tests, the Mann–-Whitney-–Wilcoxon nonparametric test and the Student's t-test (we performed 30 runs for each algorithm), to assess if these differences in the performance of the two approaches are statistically significant or not. The results obtained for the dynamic instances in terms of the total distance objective, indicate statistically significant differences on all DVRPTW instances, excepting \emph{R202-1.0}, \emph{RC102-0.5} and \emph{RC203-1.0}. Also when considering the number of vehicles objective, statistically significant differences are recorded on all dynamic instances, except for \emph{R202-0.1} and \emph{RC102-0.5}. On static instances, only for \emph{C102-0.0} the two statistical tests do not show significant differences, in terms of the total distance objective. For the number of vehicles objective, the results indicate statistically significant differences on all static instances.

Correlating the values obtained for the considered measures by the two investigated approaches, it can be stated that our algorithm achieves the best overall performance on both objectives.
Moreover, in some cases where \emph{DVRPTW-ACS} does not have such a good performance, the obtained values are close to the ones attained by \emph{MACS-DVRPTW}. This indicates that our approach is competitive with the \emph{MACS-DVRPTW} solver, and on higher dynamicity instances \emph{DVRPTW-ACS} is able to yield better results.

The good performance of our approach  can be ascribed to its underlying joint mechanism employed in the transition step of the ACS based algorithm, in which the vehicles' tours are constructed simultaneously. This offers better chances of covering the available unvisited customers with the existing vehicles, without needing to resort to additional vehicles. In contrast to this, in \emph{MACS-DVRPTW} the tours are constructed sequentially and are filled up one by one with customers. This aspect restricts the number of choices for adding an unrouted customer, which may lead to using more vehicles. Besides this, \emph{DVRPTW-ACS} is coupled with a powerful insertion heuristic, being among the best heuristics proposed by Solomon for VRPTW. Our insertion procedure manages to better incorporate unvisited available customers in the tours of existing vehicles, by searching for each unvisited node the best insertion position. In this way, all the unvisited nodes have equal chances and compete for being inserted in best position within a vehicle's tour. In contrast to this, in \emph{MACS-DVRPTW} the nodes are inserted sequentially into their best position, according to a partial order of the customers, which are sorted in descending order based on their demand.

In terms of the computational cost, the comparison of the number of feasible solutions obtained by the two approaches, reveals that for most of the DVRPTW instances, \emph{MACS-DVRPTW} computes more feasible solution than our algorithm. This may be attributed to the fact that \emph{DVRPTW-ACS} was implemented in Java, whilst the \emph{MACS-DVRPTW} solver was coded in C, which may induce more rapid execution times. Also, the relocation and exchange operators from our local search procedure are applied multiple times, and this involves a higher computational cost, compared to the cross exchange operator from \emph{MACS-DVRPTW}, which is called only once. This aspect indicates that the better performance of our algorithm on DVRPTW instances with higher dynamicity is not due to a greater number of feasible solutions that it computes.

\begin{table}
\caption{The values obtained by \emph{MACS-DVRPTW} and \emph{DVRPTW-ACS} for number of vehicles and total distance objectives on R1 instances with 0\%, 10\%, 50\% and 100\% dynamicity} \label{tab:R1results}
\begin{tabular}{c|c|c|c|c|c}
\hline
\hline
instance & measure & \multicolumn{2}{c|}{MACS-DVRPTW} & \multicolumn{2}{c}{DVRPTW-ACS} \\
\hline
& & NV & TD & NV & TD \\
\hline
R103-0.0 & average & \textbf{10.933} & \textbf{1087.923} & 14 & 1267.525 \\
& min & \textbf{10} & \textbf{1046.225} & 14 & 1246.521 \\
& max & \textbf{11} & \textbf{1148.068} & 14 & 1287.189 \\
& stdev & 0.254 & 23.397 & \textbf{0} & \textbf{10.78} \\
\cline{1-6}
R104-0.0 & average & \textbf{10} & \textbf{979.076} & 10.8 & 1072.18 \\
& min & \textbf{10} & \textbf{944.92} & \textbf{10} & 1037.729 \\
& max & \textbf{10} & \textbf{1007.948} & 11 & 1107.054 \\
& stdev & \textbf{0} & \textbf{15.742} & 0.407 & 16.243 \\
\cline{1-6}
R103-0.1 & average & \textbf{10.367} & \textbf{1078.205} & 14 & 1272.788 \\
& min & \textbf{10} & \textbf{1026.682} & 14 & 1248.78 \\
& max & \textbf{11} & \textbf{1135.584} & 14 & 1302.275 \\
& stdev & 0.49 & 21.565 & \textbf{0} & \textbf{14.11} \\
& increase(\%) & \textbf{0} & \textbf{-1.868} & \textbf{0} & 0.181 \\
\cline{1-6}
R104-0.1 & average & \textbf{10} & \textbf{969.424} & 10.533 & 1059.812 \\
& min & \textbf{10} & \textbf{934.079} & \textbf{10} & 1023.262 \\
& max & \textbf{10} & \textbf{996.333} & 11 & 1108.842 \\
& stdev & \textbf{0} & \textbf{16.031} & 0.507 & 17.035 \\
& increase(\%) & \textbf{0} & -1.147 & \textbf{0} & \textbf{-1.394} \\
\cline{1-6}
R103-0.5 & average & \textbf{12.367} & \textbf{1153.761} & 14 & 1301.596 \\
& min & \textbf{12} & \textbf{1113.695} & 14 & 1245.081 \\
& max & \textbf{13} & \textbf{1199.753} & 14 & 1350.338 \\
& stdev & 0.49 & \textbf{20.675} & \textbf{0} & 22.676 \\
& increase(\%) & 20 & 6.449 & \textbf{0} & \textbf{-0.116} \\
\cline{1-6}
R104-0.5 & average & \textbf{9.133} & \textbf{976.452} & 10.567 & 1097.715 \\
& min & \textbf{9} & \textbf{933.443} & 10 & 1054.123 \\
& max & \textbf{10} & \textbf{1013.922} & 11 & 1168.38 \\
& stdev & \textbf{0.346} & \textbf{24.099} & 0.504 & 25.714 \\
& increase(\%) & \textbf{-10} & \textbf{-1.215} & 0 & 1.58 \\
\cline{1-6}
R103-1.0 & average & 14.567 & 1327.262 & \textbf{14} & \textbf{1301.524} \\
& min & \textbf{14} & 1279.144 & \textbf{14} & \textbf{1267.567} \\
& max & 15 & 1390.228 & \textbf{14} & \textbf{1355.115} \\
& stdev & 0.504 & 29.899 & \textbf{0} & \textbf{18.167} \\
& increase(\%) & 40 & 22.263 & \textbf{0} & \textbf{1.688} \\
\cline{1-6}
R104-1.0 & average & \textbf{9.8} & \textbf{1035.714} & 10.533 & 1103.916 \\
& min & \textbf{9} & \textbf{955.579} & 10 & 1054.419 \\
& max & \textbf{10} & \textbf{1092.117} & 11 & 1161.368 \\
& stdev & \textbf{0.407} & 29.13 & 0.507 & \textbf{24.571} \\
& increase(\%) & \textbf{-10} & \textbf{1.128} & 0 & 1.608 \\
\cline{1-6}

\hline
\hline
\end{tabular}
\end{table}

\begin{table}
\caption{The values obtained by \emph{MACS-DVRPTW} and \emph{DVRPTW-ACS} for number of vehicles and total distance objectives on R2 instances with 0\%, 10\%, 50\% and 100\% dynamicity} \label{tab:R2results}
\begin{tabular}{c|c|c|c|c|c}
\hline
\hline
instance & measure & \multicolumn{2}{c|}{MACS-DVRPTW} & \multicolumn{2}{c}{DVRPTW-ACS} \\
\hline
& & NV & TD & NV & TD \\
\hline
R201-0.0 & average & \textbf{3.767} & \textbf{977.568} & 4.7 & 1307.182 \\
& min & \textbf{3} & \textbf{869.717} & 4 & 1239.081 \\
& max & \textbf{4} & \textbf{1154.818} & 5 & 1414.325 \\
& stdev & \textbf{0.43} & 77.168 & 0.466 & \textbf{47.917} \\
\cline{1-6}
R202-0.0 & average & \textbf{3} & \textbf{928.359} & 4 & 1209.136 \\
& min & \textbf{3} & \textbf{846.279} & 4 & 1157.297 \\
& max & \textbf{3} & \textbf{1019.08} & 4 & 1295.792 \\
& stdev & \textbf{0} & 39.238 & \textbf{0} & \textbf{27.184} \\
\cline{1-6}
R201-0.1 & average & 4.9 & \textbf{1075.4} & \textbf{4.533} & 1308.653 \\
& min & \textbf{4} & \textbf{966.55} & \textbf{4} & 1218.004 \\
& max & \textbf{5} & \textbf{1143.901} & \textbf{5} & 1433.059 \\
& stdev & \textbf{0.305} & \textbf{43.654} & 0.507 & 56.646 \\
& increase(\%) & 33.333 & 11.134 & \textbf{0} & \textbf{-1.701} \\
\cline{1-6}
R202-0.1 & average & 3.967 & \textbf{949.171} & 4 & 1196.264 \\
& min & 3 & \textbf{858.555} & 4 & 1163.795 \\
& max & 4 & \textbf{1044.958} & 4 & 1255.849 \\
& stdev & 0.183 & 46.793 & 0 & \textbf{22.327} \\
& increase(\%) & 0 & 1.451 & 0 & \textbf{0.561} \\
\cline{1-6}
R201-0.5 & average & 5.933 & \textbf{1262.015} & \textbf{4} & 1365.705 \\
& min & 5 & \textbf{1171.521} & \textbf{4} & 1323.148 \\
& max & 6 & \textbf{1341.553} & \textbf{4} & 1433.617 \\
& stdev & 0.254 & 45.987 & \textbf{0} & \textbf{25.78} \\
& increase(\%) & 66.667 & 34.701 & \textbf{0} & \textbf{6.785} \\
\cline{1-6}
R202-0.5 & average & 5.133 & \textbf{1148.596} & \textbf{4} & 1229.833 \\
& min & 5 & \textbf{1046.799} & \textbf{4} & 1166.377 \\
& max & 6 & \textbf{1233.396} & \textbf{4} & 1349.382 \\
& stdev & 0.346 & 48.095 & \textbf{0} & \textbf{34.816} \\
& increase(\%) & 66.667 & 23.694 & \textbf{0} & \textbf{0.785} \\
\cline{1-6}
R201-1.0 & average & 7.2 & 1811.948 & \textbf{4} & \textbf{1394.533} \\
& min & 6 & 1625.431 & \textbf{4} & \textbf{1357.513} \\
& max & 8 & 1995.438 & \textbf{4} & \textbf{1450.423} \\
& stdev & 0.484 & 86.74 & \textbf{0} & \textbf{23.731} \\
& increase(\%) & 100 & 86.892 & \textbf{0} & \textbf{9.558} \\
\cline{1-6}
R202-1.0 & average & 5.433 & 1249.129 & \textbf{4} & 1247.435 \\
& min & 5 & 1129.508 & \textbf{4} & 1195.321 \\
& max & 6 & 1415.06 & \textbf{4} & 1323.849 \\
& stdev & 0.504 & 54.7 & \textbf{0} & 33.132 \\
& increase(\%) & 66.667 & 33.468 & \textbf{0} & 3.286 \\
\cline{1-6}

\hline
\hline
\end{tabular}
\end{table}

\begin{table}
\caption{The values obtained by \emph{MACS-DVRPTW} and \emph{DVRPTW-ACS} for number of vehicles and total distance objectives on C1 instances with 0\%, 10\%, 50\% and 100\% dynamicity} \label{tab:C1results}
\begin{tabular}{c|c|c|c|c|c}
\hline
\hline
instance & measure & \multicolumn{2}{c|}{MACS-DVRPTW} & \multicolumn{2}{c}{DVRPTW-ACS} \\
\hline
& & NV & TD & NV & TD \\
\hline
C101-0.0 & average & \textbf{10} & \textbf{828.937} & \textbf{10} & \textbf{828.937} \\
& min & \textbf{10} & \textbf{828.937} & \textbf{10} & \textbf{828.937} \\
& max & \textbf{10} & \textbf{828.937} & \textbf{10} & \textbf{828.937} \\
& stdev & \textbf{0} & \textbf{0} & \textbf{0} & \textbf{0} \\
\cline{1-6}
C102-0.0 & average & \textbf{10} & 857.087 & \textbf{10} & 864.627 \\
& min & \textbf{10} & 829.133 & \textbf{10} & 834.639 \\
& max & \textbf{10} & 899.302 & \textbf{10} & 902.735 \\
& stdev & \textbf{0} & 19.344 & \textbf{0} & 19.69 \\
\cline{1-6}
C101-0.1 & average & 11 & 1003.766 & \textbf{10} & \textbf{828.937} \\
& min & 11 & 1002.601 & \textbf{10} & \textbf{828.937} \\
& max & 11 & 1003.904 & \textbf{10} & \textbf{828.937} \\
& stdev & \textbf{0} & 0.397 & \textbf{0} & \textbf{0} \\
& increase(\%) & 10 & 20.95 & \textbf{0} & \textbf{0} \\
\cline{1-6}
C102-0.1 & average & 12 & 974.14 & \textbf{10} & \textbf{845.559} \\
& min & 12 & 916.369 & \textbf{10} & \textbf{830.814} \\
& max & 12 & 1087.706 & \textbf{10} & \textbf{890.275} \\
& stdev & \textbf{0} & 38.088 & \textbf{0} & \textbf{12.499} \\
& increase(\%) & 20 & 10.521 & \textbf{0} & \textbf{-0.458} \\
\cline{1-6}
C101-0.5 & average & 16.667 & 1392.599 & \textbf{10} & \textbf{828.937} \\
& min & 16 & 1293.749 & \textbf{10} & \textbf{828.937} \\
& max & 18 & 1629.158 & \textbf{10} & \textbf{828.937} \\
& stdev & 0.661 & 68.256 & \textbf{0} & \textbf{0} \\
& increase(\%) & 60 & 56.073 & \textbf{0} & \textbf{0} \\
\cline{1-6}
C102-0.5 & average & 11.633 & 1170.861 & \textbf{10} & \textbf{868.538} \\
& min & 11 & 1040.89 & \textbf{10} & \textbf{828.937} \\
& max & 12 & 1273.721 & \textbf{10} & \textbf{932.31} \\
& stdev & 0.49 & 61.062 & \textbf{0} & \textbf{25.149} \\
& increase(\%) & 10 & 25.54 & \textbf{0} & \textbf{-0.683} \\
\cline{1-6}
C101-1.0 & average & 20.167 & 2016.359 & \textbf{10} & \textbf{828.937} \\
& min & 19 & 1879.521 & \textbf{10} & \textbf{828.937} \\
& max & 21 & 2185.256 & \textbf{10} & \textbf{828.937} \\
& stdev & 0.461 & 69.944 & \textbf{0} & \textbf{0} \\
& increase(\%) & 90 & 126.739 & \textbf{0} & \textbf{0} \\
\cline{1-6}
C102-1.0 & average & 16.833 & 1521.483 & \textbf{10} & \textbf{881.348} \\
& min & 16 & 1457.996 & \textbf{10} & \textbf{872.031} \\
& max & 17 & 1596.63 & \textbf{10} & \textbf{888.646} \\
& stdev & 0.379 & 36.923 & \textbf{0} & \textbf{5.455} \\
& increase(\%) & 60 & 75.846 & \textbf{0} & \textbf{4.48} \\
\cline{1-6}

\hline
\hline
\end{tabular}
\end{table}

\begin{table}
\caption{The values obtained by \emph{MACS-DVRPTW} and \emph{DVRPTW-ACS} for number of vehicles and total distance objectives on C2 instances with 0\%, 10\%, 50\% and 100\% dynamicity} \label{tab:C2results}
\begin{tabular}{c|c|c|c|c|c}
\hline
\hline
instance & measure & \multicolumn{2}{c|}{MACS-DVRPTW} & \multicolumn{2}{c}{DVRPTW-ACS} \\
\hline
& & NV & TD & NV & TD \\
\hline
C201-0.0 & average & 3.767 & 619.148 & \textbf{3} & \textbf{591.557} \\
& min & \textbf{3} & \textbf{591.557} & \textbf{3} & \textbf{591.557} \\
& max & 4 & 675.307 & \textbf{3} & \textbf{591.557} \\
& stdev & 0.43 & 20.922 & \textbf{0} & \textbf{0} \\
\cline{1-6}
C202-0.0 & average & 3.267 & 670.613 & \textbf{3} & \textbf{600.733} \\
& min & \textbf{3} & 605.298 & \textbf{3} & \textbf{591.557} \\
& max & 4 & 734.512 & \textbf{3} & \textbf{634.5} \\
& stdev & 0.45 & 37.403 & \textbf{0} & \textbf{15.079} \\
\cline{1-6}
C201-0.1 & average & 4 & 626.003 & \textbf{3} & \textbf{591.557} \\
& min & 4 & 622.535 & \textbf{3} & \textbf{591.557} \\
& max & 4 & 650.237 & \textbf{3} & \textbf{591.557} \\
& stdev & \textbf{0} & 9.014 & \textbf{0} & \textbf{0} \\
& increase(\%) & 33.333 & 5.237 & \textbf{0} & \textbf{0} \\
\cline{1-6}
C202-0.1 & average & 4.033 & 689.57 & \textbf{3} & \textbf{594.185} \\
& min & 4 & 626.809 & \textbf{3} & \textbf{591.557} \\
& max & 5 & 797.32 & \textbf{3} & \textbf{623.531} \\
& stdev & 0.183 & 49.843 & \textbf{0} & \textbf{7.969} \\
& increase(\%) & 33.333 & 3.554 & \textbf{0} & \textbf{0} \\
\cline{1-6}
C201-0.5 & average & 6.133 & 899.03 & \textbf{3} & \textbf{591.557} \\
& min & 6 & 814.258 & \textbf{3} & \textbf{591.557} \\
& max & 7 & 1122.569 & \textbf{3} & \textbf{591.557} \\
& stdev & 0.346 & 82.495 & \textbf{0} & \textbf{0} \\
& increase(\%) & 100 & 37.647 & \textbf{0} & \textbf{0} \\
\cline{1-6}
C202-0.5 & average & 5.933 & 878.535 & \textbf{3} & \textbf{606.853} \\
& min & 5 & 767.05 & \textbf{3} & \textbf{591.557} \\
& max & 6 & 1008.27 & \textbf{3} & \textbf{636.705} \\
& stdev & 0.254 & 75.38 & \textbf{0} & \textbf{17.391} \\
& increase(\%) & 66.667 & 26.723 & \textbf{0} & \textbf{0} \\
\cline{1-6}
C201-1.0 & average & 7.233 & 1577.455 & \textbf{3} & \textbf{591.557} \\
& min & 7 & 1339.839 & \textbf{3} & \textbf{591.557} \\
& max & 8 & 1831.838 & \textbf{3} & \textbf{591.557} \\
& stdev & 0.43 & 116.901 & \textbf{0} & \textbf{0} \\
& increase(\%) & 133.333 & 126.494 & \textbf{0} & \textbf{0} \\
\cline{1-6}
C202-1.0 & average & 6.067 & 948.312 & \textbf{3} & \textbf{593.19} \\
& min & 6 & 856.878 & \textbf{3} & \textbf{591.557} \\
& max & 7 & 1126.274 & \textbf{3} & \textbf{626.741} \\
& stdev & 0.254 & 77.829 & \textbf{0} & \textbf{6.422} \\
& increase(\%) & 100 & 41.563 & \textbf{0} & \textbf{0} \\
\cline{1-6}

\hline
\hline
\end{tabular}
\end{table}

\begin{table}
\caption{The values obtained by \emph{MACS-DVRPTW} and \emph{DVRPTW-ACS} for number of vehicles and total distance objectives on RC1 instances with 0\%, 10\%, 50\% and 100\% dynamicity} \label{tab:RC1results}
\begin{tabular}{c|c|c|c|c|c}
\hline
\hline
instance & measure & \multicolumn{2}{c|}{MACS-DVRPTW} & \multicolumn{2}{c}{DVRPTW-ACS} \\
\hline
& & NV & TD & NV & TD \\
\hline
RC101-0.0 & average & \textbf{13.567} & \textbf{1419.259} & 15.1 & 1669.585 \\
& min & \textbf{13} & \textbf{1353.045} & 15 & 1649.803 \\
& max & \textbf{14} & \textbf{1514.738} & 16 & 1705.964 \\
& stdev & 0.504 & 36.447 & \textbf{0.305} & \textbf{14.753} \\
\cline{1-6}
RC102-0.0 & average & \textbf{11.967} & \textbf{1316.636} & 13.767 & 1538.879 \\
& min & \textbf{11} & \textbf{1278.533} & 13 & 1501.709 \\
& max & \textbf{12} & \textbf{1383.166} & 14 & 1589.491 \\
& stdev & \textbf{0.183} & 24.018 & 0.43 & \textbf{21.2} \\
\cline{1-6}
RC101-0.1 & average & \textbf{14.9} & \textbf{1499.454} & 15.267 & 1698.417 \\
& min & \textbf{14} & \textbf{1438.998} & 15 & 1666.823 \\
& max & \textbf{15} & \textbf{1569.396} & 16 & 1740.601 \\
& stdev & \textbf{0.305} & 33.063 & 0.45 & \textbf{19.908} \\
& increase(\%) & 7.692 & 6.353 & \textbf{0} & \textbf{1.032} \\
\cline{1-6}
RC102-0.1 & average & \textbf{11.167} & \textbf{1361.773} & 13.667 & 1563.956 \\
& min & \textbf{11} & \textbf{1279.555} & 13 & 1514.923 \\
& max & \textbf{12} & \textbf{1460.245} & 14 & 1629.863 \\
& stdev & \textbf{0.379} & 46.071 & 0.479 & \textbf{22.929} \\
& increase(\%) & \textbf{0} & \textbf{0.08} & \textbf{0} & 0.88 \\
\cline{1-6}
RC101-0.5 & average & 18.733 & 1947.191 & \textbf{15.133} & \textbf{1733.718} \\
& min & 17 & 1772.705 & \textbf{14} & \textbf{1669.847} \\
& max & 20 & 2124.249 & \textbf{16} & \textbf{1781.071} \\
& stdev & 0.828 & 85.384 & \textbf{0.434} & \textbf{28.038} \\
& increase(\%) & 30.769 & 31.016 & \textbf{-6.667} & \textbf{1.215} \\
\cline{1-6}
RC102-0.5 & average & 14 & 1566.836 & 13.933 & 1580.218 \\
& min & 13 & 1471.19 & 13 & 1540.802 \\
& max & 15 & 1693.733 & 15 & 1651.335 \\
& stdev & 0.587 & 54.201 & 0.365 & 28.635 \\
& increase(\%) & 18.182 & 15.069 & 0 & 2.603 \\
\cline{1-6}
RC101-1.0 & average & 20.867 & 2219.684 & \textbf{15.1} & \textbf{1720.723} \\
& min & 19 & 2057.832 & \textbf{15} & \textbf{1669.856} \\
& max & 23 & 2453.04 & \textbf{16} & \textbf{1781.512} \\
& stdev & 0.819 & 86.721 & \textbf{0.305} & \textbf{26.007} \\
& increase(\%) & 46.154 & 52.089 & \textbf{0} & \textbf{1.215} \\
\cline{1-6}
RC102-1.0 & average & 16.433 & 1714.181 & \textbf{13.9} & \textbf{1602.576} \\
& min & 15 & 1613.665 & \textbf{13} & \textbf{1551.439} \\
& max & 17 & 1778.713 & \textbf{15} & \textbf{1704.936} \\
& stdev & 0.626 & 44.416 & \textbf{0.481} & \textbf{32.527} \\
& increase(\%) & 36.364 & 26.212 & \textbf{0} & \textbf{3.312} \\
\cline{1-6}

\hline
\hline
\end{tabular}
\end{table}

\begin{table}
\caption{The values obtained by \emph{MACS-DVRPTW} and \emph{DVRPTW-ACS} for number of vehicles and total distance objectives on RC2 instances with 0\%, 10\%, 50\% and 100\% dynamicity} \label{tab:RC2results}
\begin{tabular}{c|c|c|c|c|c}
\hline
\hline
instance & measure & \multicolumn{2}{c|}{MACS-DVRPTW} & \multicolumn{2}{c}{DVRPTW-ACS} \\
\hline
& & NV & TD & NV & TD \\
\hline
RC202-0.0 & average & \textbf{3} & \textbf{1004.369} & 4 & 1301.551 \\
& min & \textbf{3} & \textbf{893.646} & 4 & 1245.37 \\
& max & \textbf{3} & \textbf{1134.003} & 4 & 1366.973 \\
& stdev & \textbf{0} & 51.035 & \textbf{0} & \textbf{32.926} \\
\cline{1-6}
RC203-0.0 & average & \textbf{3} & \textbf{900.178} & 4 & 1046.39 \\
& min & \textbf{3} & \textbf{822.186} & 4 & 1011.552 \\
& max & \textbf{3} & \textbf{975.809} & 4 & 1112.606 \\
& stdev & \textbf{0} & 35.982 & \textbf{0} & \textbf{24.857} \\
\cline{1-6}
RC202-0.1 & average & \textbf{4} & \textbf{1044.917} & \textbf{4} & 1291.909 \\
& min & \textbf{4} & \textbf{952.876} & \textbf{4} & 1234.551 \\
& max & \textbf{4} & \textbf{1139.719} & \textbf{4} & 1343.794 \\
& stdev & \textbf{0} & 48.588 & \textbf{0} & \textbf{29.813} \\
& increase(\%) & 33 & 6.628 & \textbf{0} & \textbf{-0.869} \\
\cline{1-6}
RC203-0.1 & average & \textbf{3} & \textbf{899.799} & 3.967 & 1076.276 \\
& min & \textbf{3} & \textbf{814.817} & \textbf{3} & 1019.385 \\
& max & \textbf{3} & \textbf{997.529} & 4 & 1216.303 \\
& stdev & \textbf{0} & \textbf{39.368} & 0.183 & 41.685 \\
& increase(\%) & 0 & \textbf{-0.896} & \textbf{-25} & 0.774 \\
\cline{1-6}
RC202-0.5 & average & 4.833 & \textbf{1261.691} & \textbf{4} & 1340.959 \\
& min & \textbf{4} & \textbf{1146.6} & \textbf{4} & 1246.376 \\
& max & 5 & \textbf{1438.765} & \textbf{4} & 1440.575 \\
& stdev & 0.379 & 71.85 & \textbf{0} & \textbf{47.483} \\
& increase(\%) & 33.333 & 28.306 & \textbf{0} & \textbf{0.081} \\
\cline{1-6}
RC203-0.5 & average & 4.333 & \textbf{1123.38} & \textbf{3.1} & 1247.353 \\
& min & 4 & \textbf{1016.05} & \textbf{3} & 1122.897 \\
& max & 5 & \textbf{1235.297} & \textbf{4} & 1392.911 \\
& stdev & 0.479 & \textbf{50.062} & \textbf{0.305} & 67.217 \\
& increase(\%) & 33.333 & 23.579 & \textbf{-25} & \textbf{11.007} \\
\cline{1-6}
RC202-1.0 & average & 6.7 & 1688.338 & \textbf{3.967} & \textbf{1364.969} \\
& min & 6 & 1460.384 & \textbf{3} & \textbf{1277.912} \\
& max & 7 & 1825.705 & \textbf{4} & \textbf{1544.412} \\
& stdev & 0.466 & 76.98 & \textbf{0.183} & \textbf{60.915} \\
& increase(\%) & 100 & 63.419 & \textbf{-25} & \textbf{2.613} \\
\cline{1-6}
RC203-1.0 & average & 5.8 & 1231.181 & \textbf{3.067} & 1243.447 \\
& min & 5 & 1139.477 & \textbf{3} & 1097.193 \\
& max & 6 & 1334.584 & \textbf{4} & 1345.215 \\
& stdev & 0.407 & 41.599 & \textbf{0.254} & 60.904 \\
& increase(\%) & 66.667 & 38.591 & \textbf{-25} & 8.466 \\
\cline{1-6}

\hline
\hline
\end{tabular}
\end{table}

%

\section{Conclusions} \label{sec:conclusions}
In this paper, we have proposed an ACS based algorithm to tackle the single-objective DVRPTW problem, which considers a hierarchy among the two objectives, minimizing the number of vehicles having priority over minimizing the traveled distance.
Experiments conducted on several DVRPTW instances with different dynamicity levels, showed that our algorithm is efficient and competitive with another ACO based approach from the literature. Besides that, on DVRPTW instances with a higher dynamicity, \emph{DVRPTW-ACS} was able to achieve better results than \emph{MACS-DVRPTW}, in terms of both objectives. Also, our algorithm proved to cope better on higher dynamicity instances, since the decline in the solution quality is not so significant, compared to the solution obtained for the corresponding static instances.
This makes our approach more suitable to be applied on DVRPTW instances with a higher dynamicity, in which few customers requests are a priori known, case more likely to occur in real-world scenarios.

Possible avenues for future work include solving this problem from a multi-objective perspective and extending this research to tackle dynamic rich vehicle routing problems, that can incorporate more complex constraints and objectives, found in real-life VRPs.

\section*{Acknowledgements}
We would like to thank to the authors of \emph{MACS-DVRPTW} solver \cite{Veen2013} for providing us their source code, that allowed us to perform the comparison with our algorithm.
\begingroup
\setlength\intextsep{0pt}
\begin{wrapfigure}[2]{r}{0.10\textwidth}
\includegraphics[width=\linewidth]{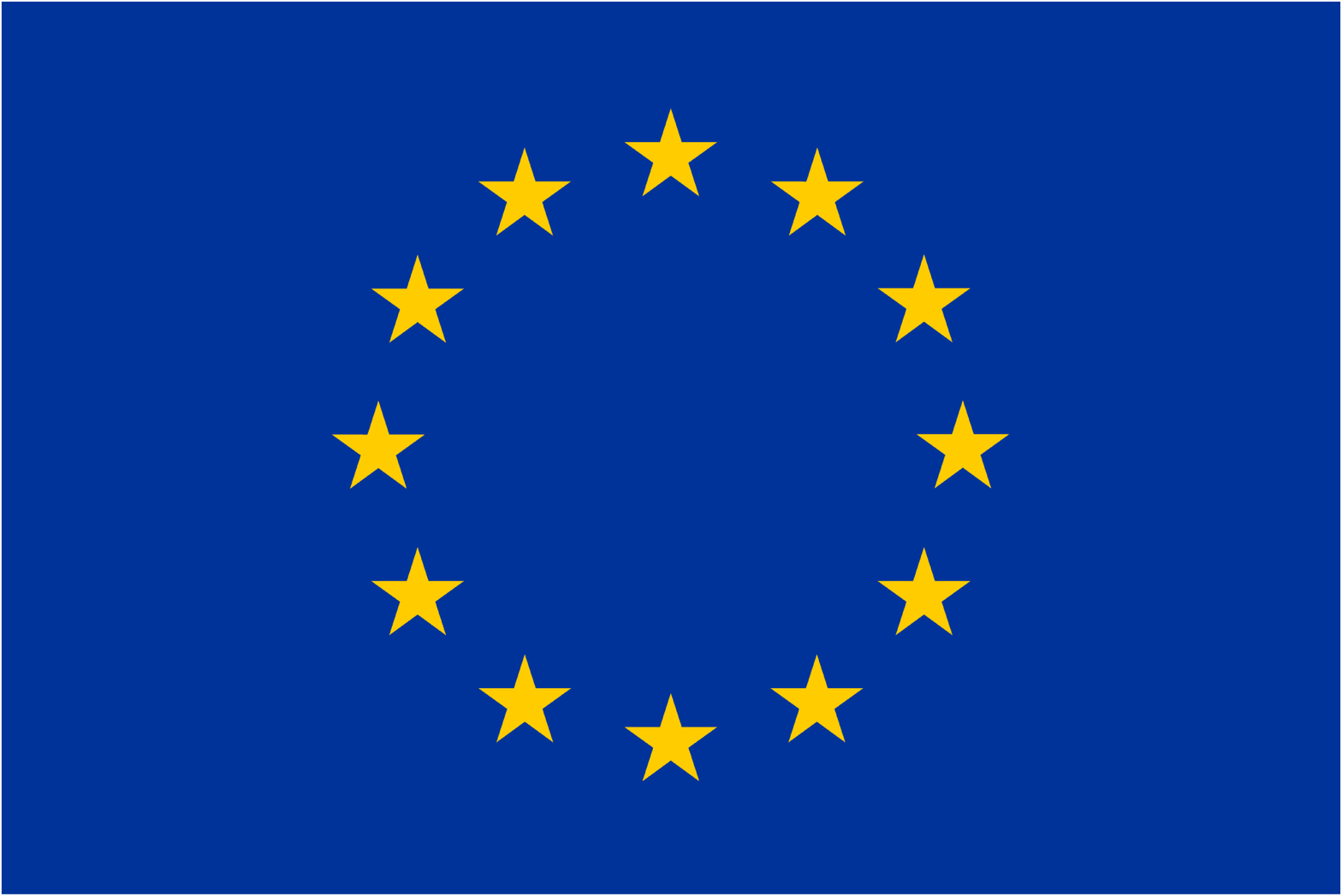}
\end{wrapfigure}
This work is partly funded from the \emph{European Union{\textquotesingle}s Horizon 2020 research and innovation programme} under grant agreement No 692178. \par
\endgroup


\end{document}